\documentclass[runningheads]{llncs}
\usepackage[pdftex]{graphicx}
\usepackage{graphicx} 
\usepackage{subcaption}

\usepackage{cite}
\usepackage{amsmath,amssymb,amsfonts}
\usepackage{algorithmic}
\usepackage{textcomp}
\usepackage{xcolor}

\AtBeginDocument{%
  \providecommand\BibTeX{{%
    \normalfont B\kern-0.5em{\scshape i\kern-0.25em b}\kern-0.8em\TeX}}}

\author{Tran Hien Van\inst{4} \and Abhay Goyal\inst{1}  \and Muhammad Siddique\inst{2}  \and Lam Yin Cheung\inst{2} \and Nimay Parekh\inst{7} \and Jonathan Y	Huang \inst{3} \and Keri McCrickerd \inst{3} \and Edson C	Tandoc Jr.\inst{4} \and Gerard	Chung\inst{5} \and Navin Kumar\inst{6} }

\institute{Missouri S \& T, Missouri MO 65401, USA,\\
\email{aghnw@umsystem.edu}
\and
Yale University, Yale, CO,  USA \\ 
\email{\{siddique9171, yclam\} @gmail.com} 
\and
Singapore Institute for Clinical Sciences \\
\email{jonathan\_huang@sics.a-star.edu.sg}
\and
Nanyang Technical University \\ 
\email{hienvantran1510@gmail.com, edson@ntu.edu.sg}
\and
National University of Singapore \\ 
\email{gerard@nus.edu.sg}
\and
New York University, NY, USA \\
\email{navin183@gmail.com}
\and
Jivox, USA\\
\email{nrparekh@gmail.com}
}

\begin{document}


\title{How is Fatherhood Framed Online in Singapore?}

\maketitle

\begin{abstract}
The proliferation of discussion about fatherhood in Singapore attests to its significance, indicating the need for an exploration of how fatherhood is framed, aiding policy-making around fatherhood in Singapore. Sound and holistic policy around fatherhood in Singapore may reduce stigma and apprehension around being a parent, critical to improving the nation's flagging birth rate. We analyzed 15,705 articles and 56,221 posts to study how fatherhood is framed in Singapore across a range of online platforms (news outlets, parenting forums, Twitter). We used NLP techniques to understand these differences. While fatherhood was framed in a range of ways on the Singaporean online environment, it did not seem that fathers were framed as central to the Singaporean family unit. A strength of our work is how the different techniques we have applied validate each other.

\end{abstract}

\section{Introduction}
Fatherhood is now an unprecedentedly visible cultural phenomenon in Singapore. This increased attention is related to the inaugural nationwide fatherhood movement, Dads for Life, the continual development of parenting magazines and the recent emergence of fatherhood blogs within the Singapore internet sphere. In recent times, various fatherhood-related initiatives in Singapore have collaborated with government agencies, business corporations, and community organizations on initiatives to create awareness of the importance of the father’s role, develop commitment to good fathering, and encourage fathers to spend time with their children. In Singapore, the introduction of paternity leave and encouragement for fathers to play a bigger role in childcare and child-raising suggest that the government is sympathetic to the pursuit of gender equality. However, there is a gap between the perception of the importance of fathers and the actual involvement of fathers in their children’s lives. In addition, the role of fathers continues to be recognized primarily as that of a breadwinner. Yet fathers want to do more and experience parenthood as a very fulfilling experience, to which they are highly committed \cite{lim2021confucian}. The proliferation of discussion about fatherhood in Singapore attests to its significance as a commercial, ideological, and cultural subject, indicating the need for an exploration of how fatherhood is framed, aiding policy-making around fatherhood in Singapore. While there has been research around how fatherhood is framed in the Singapore context, there is limited analysis of how fatherhood is framed on social media, news outlets, or online forums. Such platforms are where opinions or news on fatherhood are forwarded, people get parenting information, or get quick answers to fatherhood questions. Studying how fatherhood is framed in the online Singaporean context is central to crafting progressive and effective policy around parenting in Singapore, as well as managing the media landscape. Sound and holistic policy around fatherhood in Singapore may reduce stigma and apprehension around being a parent, critical to improving the nation's flagging birth rate. Policies developed in Singapore around fatherhood may then be implemented in nearby East Asian countries, which have similarly low birth rates, to mitigate a rapidly aging society and a shrinking taxpayer base. In this paper, we demonstrate how fatherhood in Singapore is framed on multiple online platforms (news outlets, parenting forums, Twitter). Our main research question (RQ) is as follows: How is fatherhood in Singapore framed on various online platforms? Our findings suggested that while fatherhood was framed in a multiplicity of forms online, it did not seem that fathers were core to the family. 

\section{Related Work}
\textbf{Fatherhood Framing Online}
Work on fatherhood in Singapore is limited. Recent work proposed the concept of Confucian masculinity to explain how the depiction of active fatherhood reinforced the ubiquitous \textit{normal family} that upholds patriarchal ideology and perpetuates patriarchal power, obscuring the contradictions of class, race, and sexuality that exist in Singapore \cite{lim2021confucian}. Other work examined the fatherhood discourses in \textit{new dad} ads; feature articles from Today’s Parents, a parenting magazine; articles from Life Dads, a government electronic newsletter on fatherhood; and blog entries from three fatherhood blogs \cite{see2017most}. The study employed critical discourse analysis, and proposed a Hegemonic Fatherhood Discourse Schema to postulate that the \textit{new father/man and traditional father/man} ideology is the hegemonic fatherhood in Singapore, ultimately serving the interests of the Singapore state. While past work detailed framing around fatherhood in Singapore, previous research did not compare framing across online platforms, or provide an overview of fatherhood framing to develop policy or informational tools. While there was limited fatherhood research in the Singapore context, there was relatively more research on fatherhood framing online in other contexts. For example, recent work \cite{teague2018exploring} used discussion threads from two Web-based parenting communities, r/Daddit and r/PreDaddit from Reddit. Results demonstrated that men used web-based communities to share the joys and challenges of the fatherhood experience. 

\section{Data and Method}
\textbf{Data} We first selected three content experts who had published at least ten peer-reviewed articles in the last three years around fatherhood. We ensured the content experts were either from Singapore or conducted research on fatherhood/parenthood in Singapore. Given the wide disciplinary focus of fatherhood research, we sought to select a range of experts across disciplines. We recruited one expert from each of these disciplines: Public policy, social work, computational social science. Selecting experts from a range of fields allows results to be contextualized to fields where fatherhood research is concentrated, allowing for findings to be drawn on by stakeholders in public policy, social work, and computational social science. The context experts separately developed lists of online platforms most relevant to fatherhood in Singapore. Each expert developed a list of ten platforms independently, and we selected only platforms common to all three experts' lists. For each online platform, experts also provided up to 10 examples, where applicable, of websites, or forums, and we selected examples common to all experts' lists. The final list of platforms is as follows: Singapore news outlets (Straits Times, Channel NewsAsia, TODAYonline), parenting forums (singaporemotherhood.com, singaporeparents.com.sg/forum, forums.hardwarezone.com.sg/threads/welcome-to-hwzs-parenting-kids-early-learning-forum.5684416, mummysg.com/forums), Twitter (filtering only posts related to Singapore). Examples of platforms not selected: Facebook, Instagram, Reddit, LinkedIn. We were not able to collect Facebook and Instagram data as there was limited support for CrowdTangle, the main mode of Facebook/Instagram data collection. Similarly, the pushshift.io Reddit API had limited support and Reddit data collected was incomplete. LinkedIn had limited fatherhood posts and posts were mostly centered on non-family content. To capture fatherhood-related text on these platforms, we used queries based on a related systematic review e.g., father* OR dad* OR patern* OR paternal OR paternity OR stepdad* OR stepfather* OR step-dad* OR Step-father* OR papa. We used only English-language keywords as most of discussion in the Singapore internet environment is in English. English is also the major language of communication in Singapore. For forums, we used automated scraping techniques (Beautiful Soup) to obtain forum posts from 2010 to 2023, with the same set of keywords. We ran a search for querying the keywords in the title of the forum post or replies to the forum post. We collected all posts that contained these keywords within the forum posts and replies. Regarding Twitter, we used the Twitter API and the indicated keywords to collect tweets from 2011 to 2023. Finally, for news articles, we used Nexis to obtain news archives from 1992 to 2023. To prepare the data for analysis, English stop words such as \textit{the, a, an} were removed, along with abbreviations, and terms were stemmed using Porter’s stemming algorithm. Stemming converts words with the same stem or root (e.g., innovative and innovator) to a single word type (e.g., innovate). We organized data into four streams for analysis: Twitter (tweets), news (news articles), forums (forum posts).

\textbf{Sentiment}
Sentiment analysis can aid us in comprehending how sentiment around fatherhood is expressed in the online arena. As an example, forums may be more likely to have lower sentiment compared to news. DistilBERT was used for sentiment analysis. DistilBERT was used separately on data from each platform. The model assigns sentiment based on each article or post. Sentiment is from a -1 to 1 scale, where values $<$0 are negative sentiment, $>$0 are positive sentiment, and close to 0 are neutral. To stay within the admitted input size of the model, the text length (title + body text) was clipped to to 512 tokens.

\textbf{Emotion Recognition}
Emotion recognition can help us understand how emotions are expressed across various platforms, indicating differences in how fatherhood is framed in Singapore. For example, forums may be more likely to contain anger compared to news. We used DistilBERT for emotion recognition. The model was applied separately on data from each platform. The model assigns emotions (anger, fear, joy, love, sadness, surprise) based on each article or post. To stay within the admitted input size of the model, we clipped the length of the text (title + body text) to 512 tokens.

\begin{table}[!ht]
\centering

\begin{tabular}{|l|c|l|}
\hline
Platform & Data collected (e.g., N of posts, articles) \\ \hline
News outlets & 15,705 articles, 9,811,513 words       \\ \hline
Twitter & 54,283 tweets, 900,939 words             \\ \hline
Parenting forums  & 969 threads, 425,966 words  \\ \hline
\end{tabular}

\caption{Data collected across online platforms}
\label{tab:table_1}
\end{table}

We provided an overview of the data in Table \ref{tab:table_1}. Two reviewers independently examined 10\% of the articles or posts within each dataset to confirm salience with our research question. The reviewers then discussed their findings and highlighted items deemed relevant across both lists. We noted the following relevance proportions: News outlets (82\%), Twitter (90\%), Parenting forums (78\%). 

\section{Results}
\textbf{Overview}
We first explored sample posts across platforms. News outlets generally mentioned fatherhood in the context of providing demographic data about interviewees, with excerpts such as \textit{So the 40-year-old eye specialist and father of three had to wrap up his work at the hospital quickly}, or when interviewees were referring to their fathers with no specific reference to fatherhood e.g., \textit{Mr Lee, whose father founded the clan association, rents out its third floor to a small media firm}. Broadly, news outlets did not seem to focus on the experience of fatherhood, with the bulk of articles mentioning fathers as a demographic indicator. Twitter posts focused on people recounting incidents, often humorous or heart-warming, with their fathers e.g., \textit{My dad was telling me something serious and he hit his leg against the table and I burst out laughing so he had no choice but to laugh}, \textit{Dad brought back homemade fresh horfun (noodles) from the temple. It's delicious}. Twitter seemed to have a greater focus on fathers playing a core function in the Singapore family unit. Posts from forums were very diverse topically. Several posts were about hiring a helper for a young child: \textit{My husband is totally against the idea of employing a helper, as he does not like a stranger living with us}; \textit{I am a father of a newborn baby girl. I recently engaged a confinement lady by the name of Auntie Judy}. Such posts suggest the significant role domestic helpers play in the Singaporean family, and how a portion of a father's role is perhaps to oversee the hiring of the domestic helper. Other posts were about suspected infidelity e.g., \textit{So my Wife of 2 years has been cheating on me with another male colleague}, perhaps indicative of the strain parenting is related to within some Singaporean families. 

\begin{figure*}
    \centering
    \subfloat[News\label{fig:News}]
    {{\includegraphics[height=3cm, width=3cm]{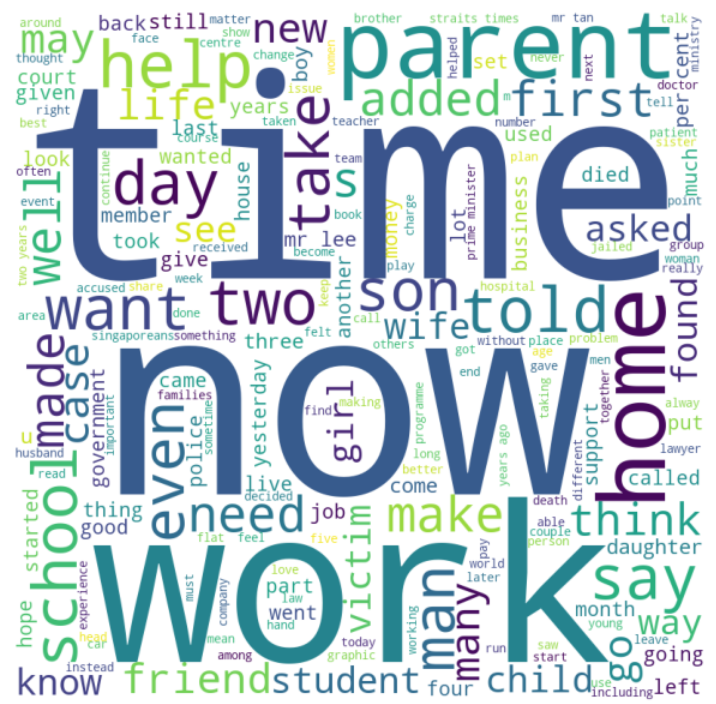}}}\qquad
    \hspace*{-0.5cm}
    \subfloat[Twitter\label{fig:Twitter}]{{\includegraphics[height=3cm, width=3cm]{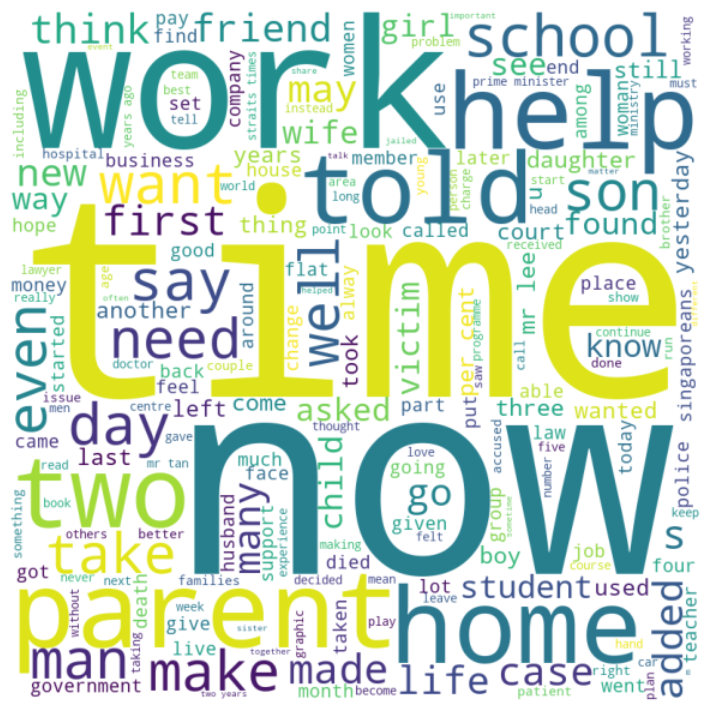}}}\qquad
    \hspace*{-0.5cm}
    \subfloat[Forums\label{fig:Forums}]{{\includegraphics[height=3cm, width=3cm]{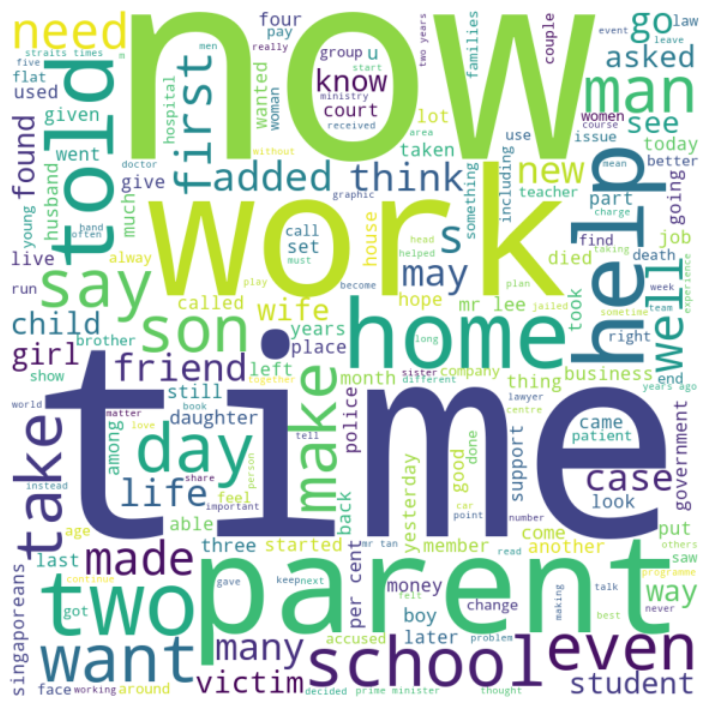}}}\qquad
\caption{Word cloud visualizations for news (\ref{fig:News}), Twitter (\ref{fig:Twitter}), forums (\ref{fig:Forums}) based on keywords relevant to fatherhood in Singapore} \label{fig:WC}
\end{figure*}


We then provided word clouds in Figure \ref{fig:WC} as an overview of the data. Across all datasets, words such as \textit{time, work, now} were prominent, perhaps indicative of how work and likely limited time are central to fatherhood in Singapore. Most common trigrams for news articles centered on leaders of Singapore, who were father and son: \textit{Lee Kwan Yew} and \textit{Lee Hsien Loong}. This may indicate that the mainstream news media discussion around fatherhood had little to do with fathers' role in a family, but simply around familial relationships within major news stories. In 1992 - 2003, common trigrams in the news were \textit{engineer success story} and \textit{pressure parent counting}. From 2004 - 2019, common trigrams were \textit{two baby boy}, \textit{first new baby}, and \textit{first time parent}. From 2020 - 2022, common trigrams were \textit{generation grit family}, and \textit{grit family love}. Broadly, news trigrams may detail how the initial focus was on children bringing pride and wealth to their families, with a transition toward celebrating new births. In more recent years, forums tended to focus on how the family unit could overcome struggles. The most common trigrams in Twitter focused on celebrating fathers through specific events such as Father's Day and birthdays: \textit{happy father's day}, \textit{happy birthday daddy}. Such phrases indicated that Twitter may be used to celebrate fathers, but only in relation to pre-defined events, instead of fathers being celebrated for time put toward caregiving etc. Common trigrams in 2011 - 2020 were \textit{love u dad}, \textit{dad love love}. 2021 onwards, popular trigrams were \textit{feel fulfilling husband}, and \textit{last nite daddy}. Twitter data demonstrated a shift from declaring love for one's father, to fathers indicating how they were fulfilled in their role. Unlike other datasets, there appears to be a shift towards a more active form of fatherhood in Singapore, where fathers describe pride in their role. Trigrams in forums centered on perceived marital infidelity, such as \textit{wife unfaithful husband}, and assisted reproductive technologies, such as \textit{ivf mommy toben}, and \textit{cousin egg donor}. Forums seemed to be platforms where people sought support around spousal infidelity and assisted reproductive technologies, rather than discuss fathers' role in the family unit. The most common trigrams in forums changed over time, with phrases such as \textit{gave birth daughter}, and \textit{first time dad} in 2010 - 2019, but with phrases such as \textit{happen file divorce}, and \textit{judged urged divorcing} in 2020. In 2021, common trigrams were \textit{conceiving single women}, while in 2022, trigrams such as \textit{crave physical intimacy}, and \textit{physicial intimacy normal} were popular. Forums, while initially around celebrating birth, may have become places where people sought information around divorce, assisted reproductive technologies, and physical intimacy. Broadly, descriptive data indicated shifting framing around fatherhood, but a limited focus on fathers as core to the Singapore family. 

\textbf{Sentiment}
\begin{table}[]
\centering

\begin{tabular}{|c|c|c|c|c|c|c|}
\hline
         & News & Twitter  & Forums \\ \hline
Positive   & 53.7\%   & 57.0\% & 27.2 \%       \\ \hline
Negative     & 43.8\% & 33.8\%  & 65.9 \%      \\ \hline
Neutral      & 2.5\% & 9.1\% & 6.9 \%    \\ \hline

\end{tabular}
\caption{Sentiment analysis breakdown for various platforms.}
\label{tab:sentiment}
\end{table} 

\begin{figure}[htbp]
     \centering
     \begin{subfigure}[b]{\textwidth}
         \centering
         \includegraphics[scale=0.3]{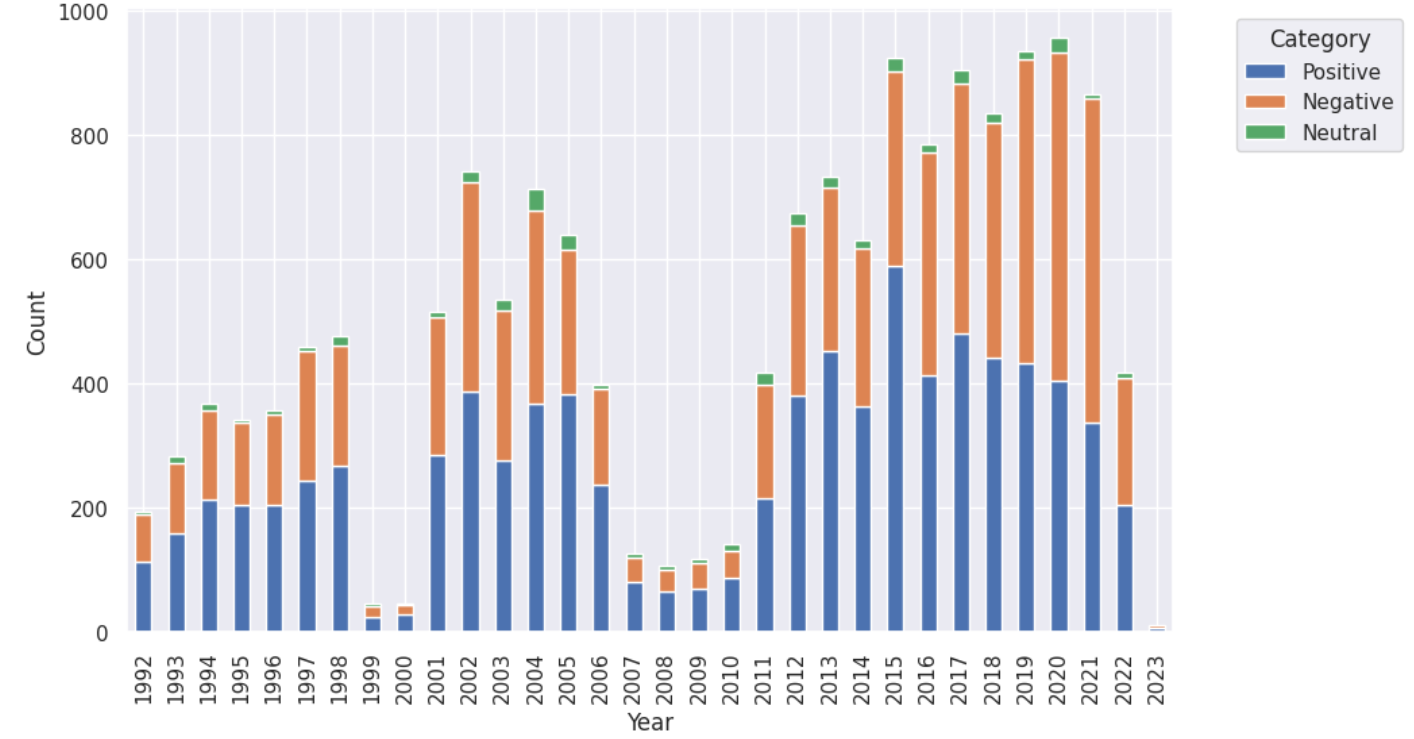}
         \caption{News}\label{fig:news}
     \end{subfigure}
     \hfill
     \begin{subfigure}[b]{\textwidth}
         \centering
         \includegraphics[scale=0.3]{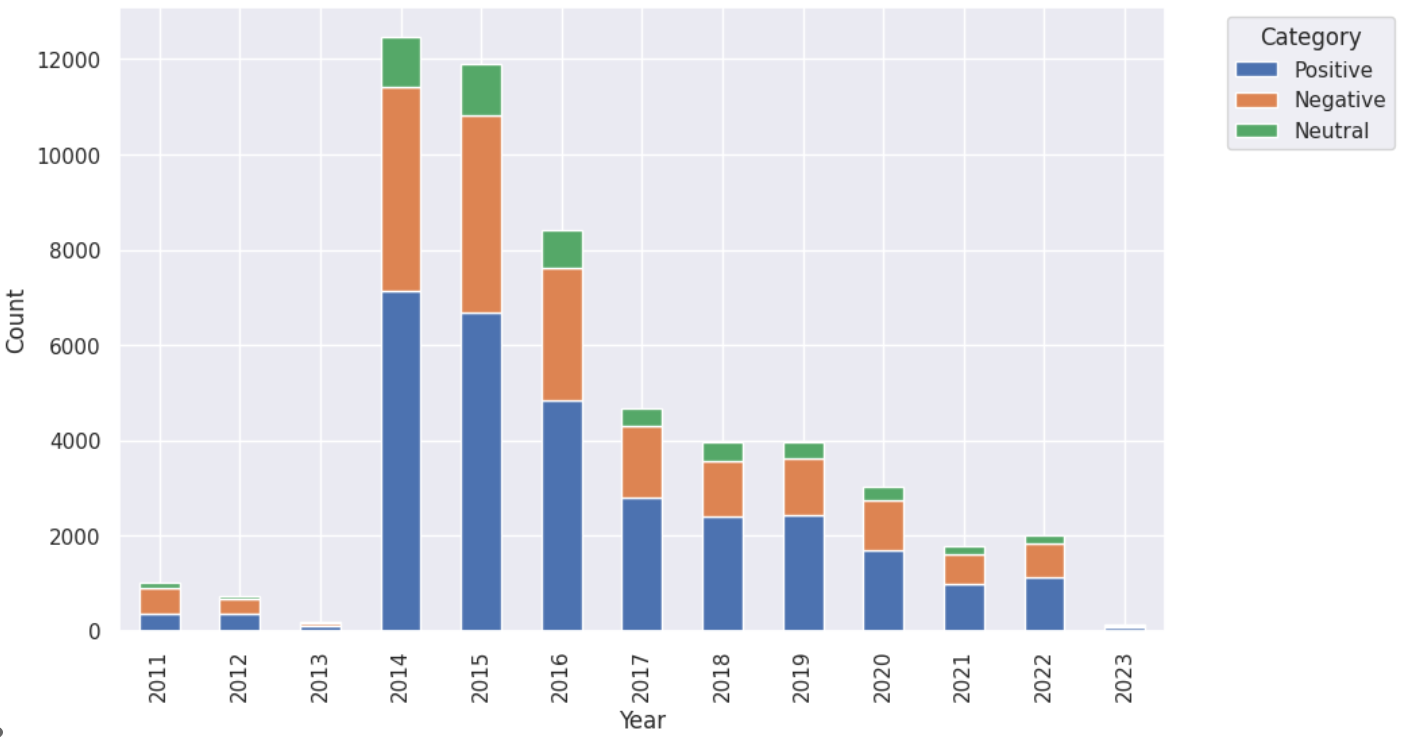}
        \caption{Twitter}\label{fig:Twitter}
     \end{subfigure}
     \hfill
     \begin{subfigure}[b]{\textwidth}
         \centering
         \includegraphics[scale=0.3]{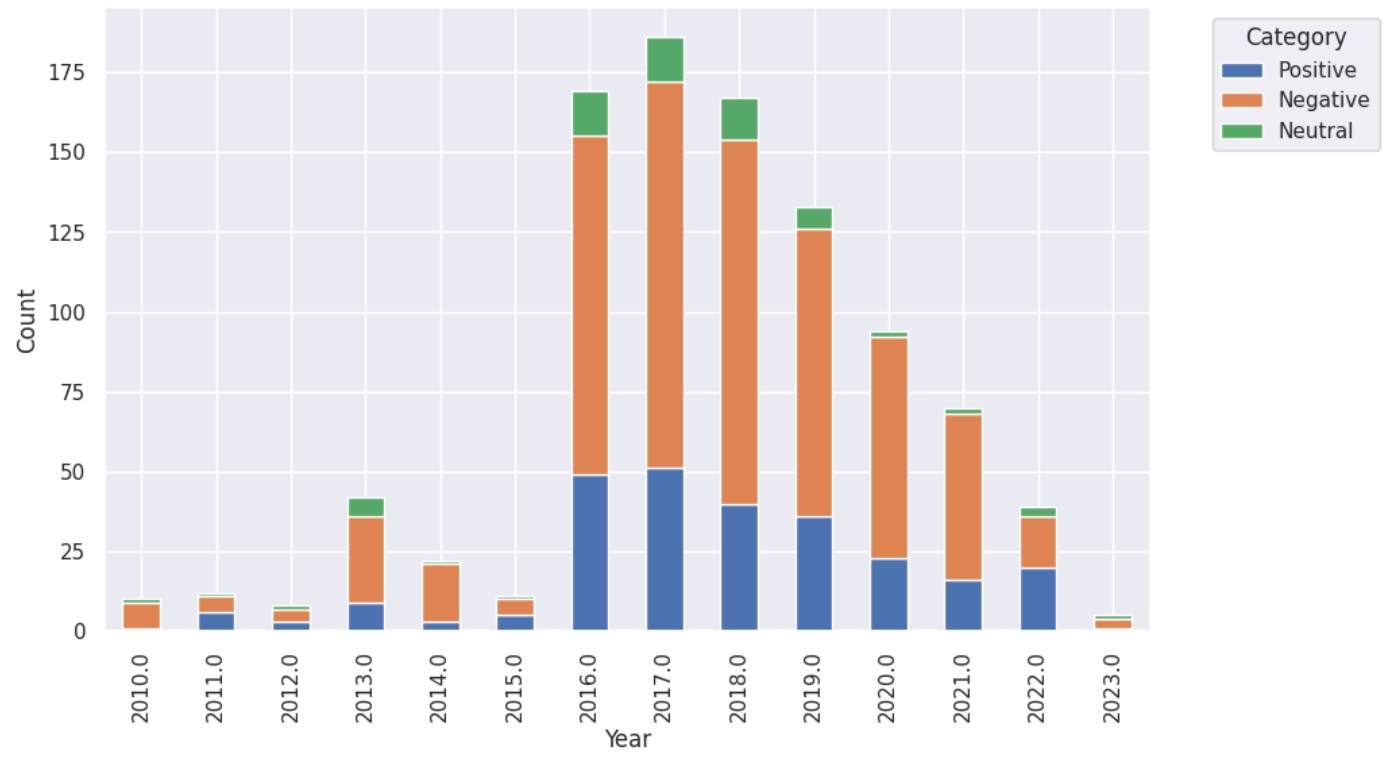}
  \caption{Forums}\label{fig:Forums}
     \end{subfigure}
        \caption{Sentiment analysis over time for various platforms.}
\label{fig:sentimentanalysis}
\end{figure}

We presented sentiment analysis results across each platform in Table \ref{tab:sentiment}. News and Twitter had higher proportions of positive sentiment (53.7\% and 57.0\% respectively) compared to forums (27.2\%). Forums had the highest proportion of negative sentiment (65.9\%), compared to news and Twitter (43.8\% and 33.8\% respectively). We then presented sentiment analysis results over time for each platform in Figure \ref{fig:sentimentanalysis}. News data exhibited several fluctuations but had the greatest rise in positive sentiment post-2009. The nationwide fatherhood movement, Dads for Life, started in 2009, may explain the increase in positive sentiment. Examples of news article content with positive sentiment were as follows: \textit{A group of prominent figures from various organisations and businesses have banded together to start up the Fathers Action Network. The network aims to kick-start a movement called Dads for Life to get fathers more involved with their families, especially in their childrens' lives. This follows a fatherhood perception survey conducted in April and May this year by a Ministry. Most felt that being a father and raising children is one of the most fulfilling experiences a man can have.}; \textit{Work is work and family is family. Our ultimate goal is still our family. Work is just a means to get the money so we should be very clear about it. And that is the sort of spirit that the Dads for Life movement wants to inspire.} After 2017, positive sentiment declined over time, and was overtaken by negative sentiment. Forums had broadly negative sentiment 2015 onward, reaching a peak in 2017, followed by a steady decline. Twitter exhibited mostly positive sentiment 2013 onward with a steady decline after. We suggest that the high proportion of positive sentiment in the news may be related to governmental initiatives and the high proportion of negative sentiment in forums may be related to a more frank discussion of the stresses of parenting. 

\textbf{Emotion Recognition}
\begin{table}[]
\centering

\begin{tabular}{|c|c|c|c|c|c|c|}
\hline
         & News & Twitter  & Forums  \\ \hline
Anger    & 32.7\%   & 28.5\% & 25.5      \\ \hline
Fear     & 18.8\% & 4.1\%  & 6.3      \\ \hline
Joy      & 61.3\% & 56.6\% & 44.2   \\ \hline
Love     & 34.2\%  & 2.4\%  & 4.1   \\ \hline
Sadness  & 2.0\%  & 7.6\% & 18.9      \\ \hline
Surprise & 11.4\%  & 0.8\%  & 1.0       \\ \hline
\end{tabular}
\caption{Emotion recognition breakdown for various platforms.}
\label{tab:emotion}
\end{table}

\begin{figure}[htbp]
     \centering
     \begin{subfigure}[b]{\textwidth}
         \centering
         \includegraphics[scale=0.3]{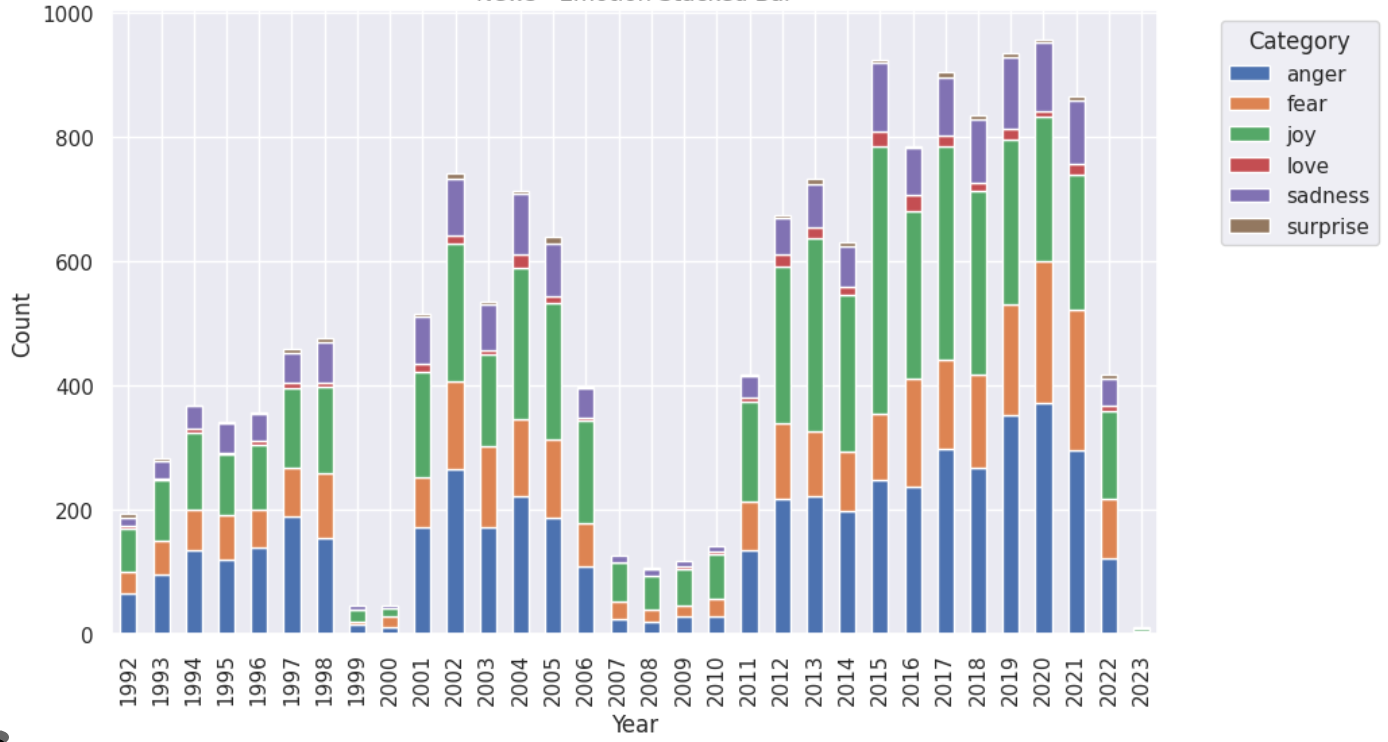}
         \caption{News}\label{fig:news}
     \end{subfigure}
     \hfill
     \begin{subfigure}[b]{\textwidth}
         \centering
         \includegraphics[scale=0.3]{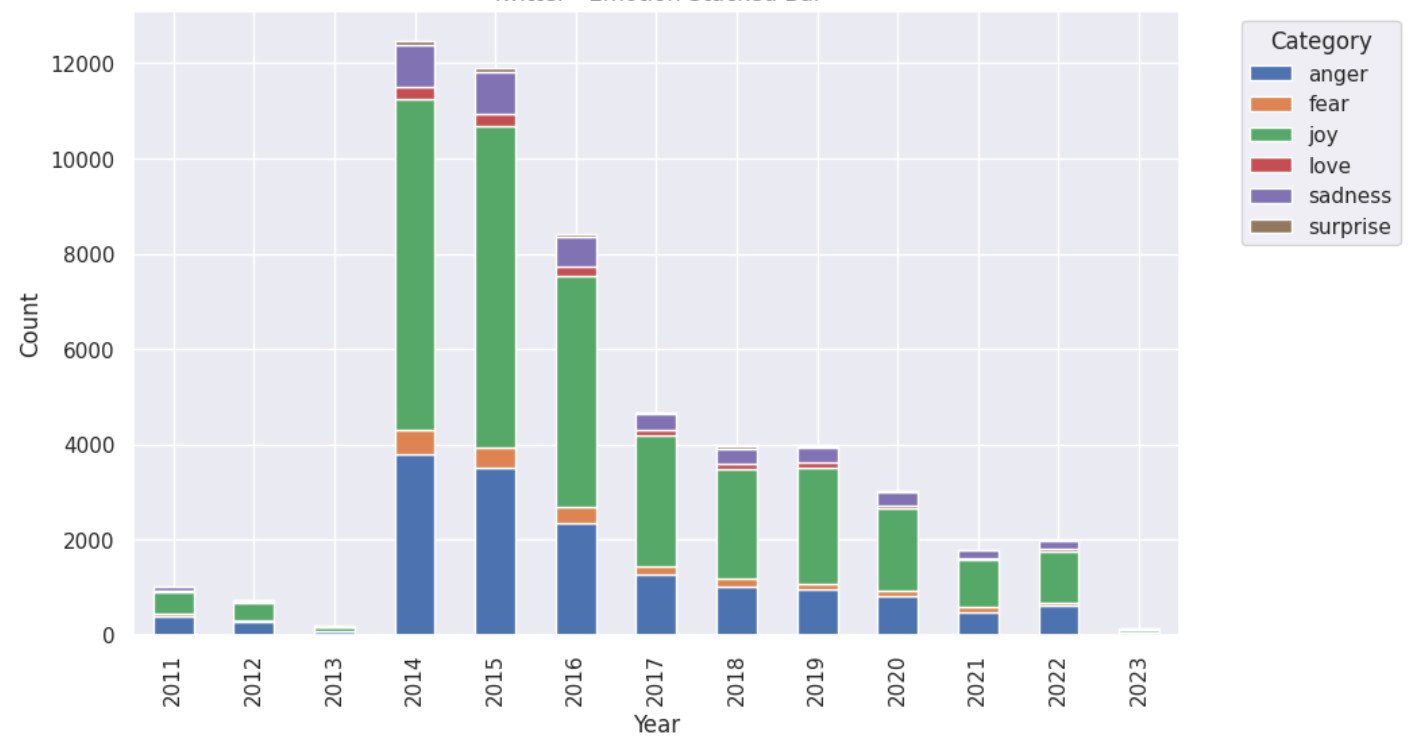}
        \caption{Twitter}\label{fig:Twitter}
     \end{subfigure}
     \hfill
     \begin{subfigure}[b]{\textwidth}
         \centering
         \includegraphics[scale=0.3]{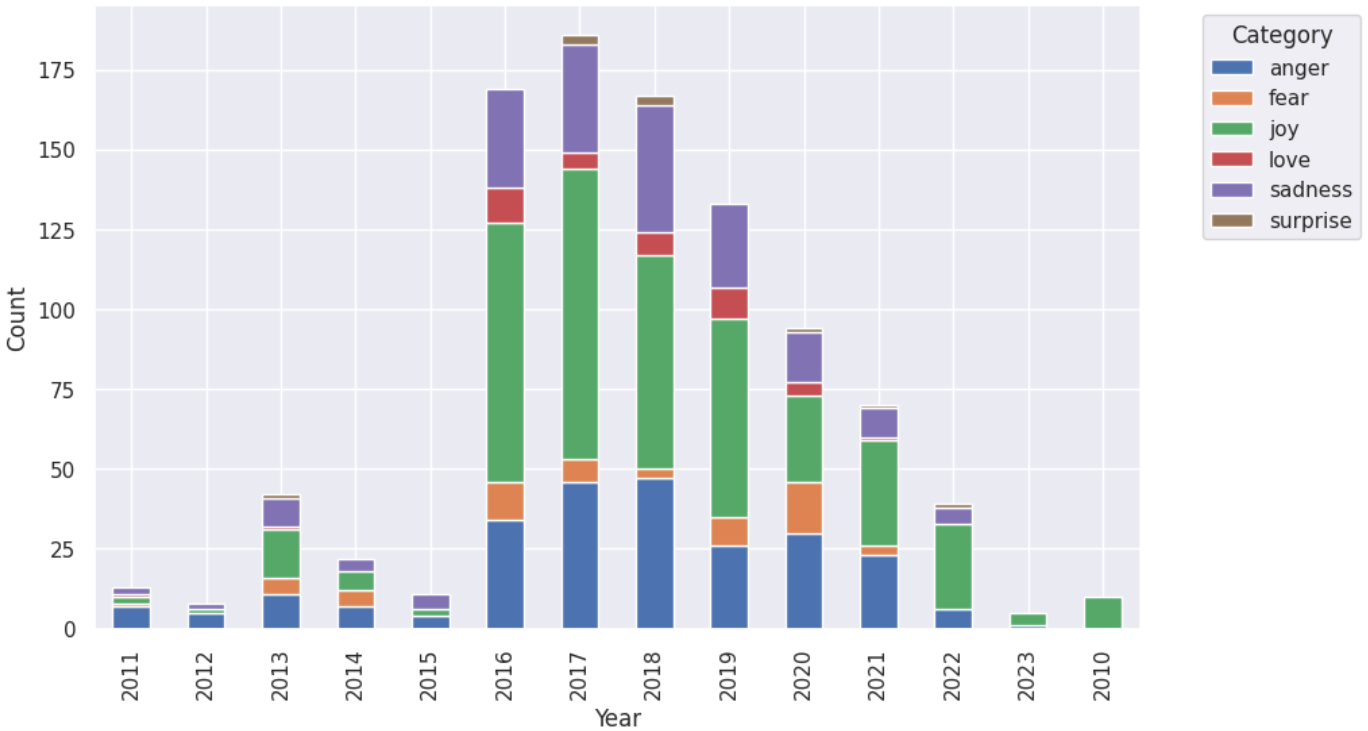}
  \caption{Forums}\label{fig:Forums}
     \end{subfigure}
        \caption{Emotion recognition over time for various platforms.}
\label{fig:emotionrec}
\end{figure}

We presented emotion recognition results across each platform in Table \ref{tab:emotion}. News had the highest proportion of joyous (61.3\%) and loving (34.2\%) posts, perhaps reflecting governmental initiatives around fatherhood. While Twitter and forums had similar levels of joyous posts (56.6\% and 44.2\% respectively), they were still not as high as news. Similarly, loving posts on Twitter and forums (2.4\% and 4.1\% respectively) were far lower than news outlets. We suggest that the emotion in the news reflects pro-fatherhood governmental initiatives, but these do not always filter successfully to other media. We then presented emotion recognition results over time for each platform in Figure \ref{fig:emotionrec}. News data exhibited several fluctuations but had the steepest rise post-2009. Dads for Life, started in 2009, may explain the uptick in news articles, especially around joy. Examples of news article content that were coded as joy: \textit{It's a happy Father's Day for SAFRA, as it is set to receive funds from the "Dads for Life" movement to pump up father-friendly activities for its members over the next two years.}; \textit{He will be running alongside his daughter in the Dads For Life 800m Father and Child Challenge, a new category in the annual SAFRA Singapore Bay Run and Army Half-Marathon. Mr Shariff, who was born without part of his left leg, said: I signed us up because I want to show her how running can make her happy.} Both Twitter and forum posts saw a sudden spike post-2013 onward, mostly around joy. We suggest that the shift in emotion may be due to a delayed reaction to Dads for Life. Broadly, we forward that the 2009 Dads for Life movement and other similar policies may have catalyzed emotional reactions around fatherhood in the Singapore online arena. However, the rises in emotion were not sustained and seemed to decline by 2023, perhaps indicative that new policy levers may need to be rolled out.  

\section{Discussion}
Our RQ was to explore how fatherhood in Singapore is framed on various online platforms. A strength of our work is how the different techniques we applied validate each other as well as reveal differences across platforms. While fatherhood was framed in a range of ways on the Singaporean online environment, it did not seem that fathers were framed as central to the Singaporean family unit. Results also indicated that governmental initiatives may have some effect on altering the framing of fatherhood, but are not lasting in effect. The concordance in our results suggests the veracity of our findings and we hope that results can add to research and policy around fatherhood in Singapore. Our evidence adds to previous research, where we provided data on how governmental initiatives may initially buttress framing around fatherhood, but needs to be sustained to provide broad and lasting support for fathers. Key to how fatherhood is framed in Singapore is the inclusion of fathers' viewpoints when writing news articles on fatherhood. Where possible, fathers themselves should be consulted on articles about fatherhood. For example, a panel staffed by fathers can comment on fatherhood-related online news articles, providing suggestions on how articles can more accurately represent fathers' concerns \cite{chen2022partisan,kumar2022online}. Our findings relied on the validity of data collected with our search terms. We used a range of established techniques to search for all articles/posts relevant to fatherhood, and our data contained text aligned with how fatherhood is framed. We were thus confident in the comprehensiveness of our data. We only used English-language text but will include other languages in future work. Given the token limits for the emotion recognition technique, we were not able to use emotion recognition for the entirety of longer news articles. We note that the recall of the search string was not tested. We note that our data may not be generalizable to how fatherhood is framed globally. Our goal was not to identify who was doing the framing around fatherhood e.g., family members or government. Future studies will seek to identify which stakeholders were likely involved in the framing. 



\bibliographystyle{splncs04}
\bibliography{ref}

\end{document}